\documentclass[10pt,twocolumn,letterpaper]{article}

\usepackage[pagenumbers]{cvpr} 

\usepackage{graphicx}
\usepackage{amsmath}
\usepackage{amssymb}
\usepackage{booktabs}
\usepackage{threeparttable}
\usepackage{multirow}
\usepackage{colortbl}
\usepackage{xcolor}
\usepackage[accsupp]{axessibility}  

\usepackage[pagebackref,breaklinks,colorlinks]{hyperref}

\usepackage[capitalize]{cleveref}
\crefname{section}{Sec.}{Secs.}
\Crefname{section}{Section}{Sections}
\Crefname{table}{Table}{Tables}
\crefname{table}{Tab.}{Tabs.}

\def\cvprPaperID{10195} 
\def\confName{CVPR}
\def\confYear{2023}

\begin{document}

\title{Self-Correctable and Adaptable Inference for \\Generalizable Human Pose Estimation}

\author{
Zhehan Kan\textsuperscript{1}, Shuoshuo Chen\textsuperscript{1}, Ce Zhang\textsuperscript{1}, Yushun Tang\textsuperscript{1}, Zhihai He\textsuperscript{1,2}\thanks{Corresponding author.}\\
{\normalsize 
\textsuperscript{1}Department of Electronic and Electrical Engineering, Southern University of Science and Technology, Shenzhen, China}\\
{\normalsize 
\textsuperscript{2}Pengcheng Laboratory, Shenzhen, China}\\
{\tt\small \{kanzh2021,chenss2021,zhangc2019,tangys2022\}@mail.sustech.edu.cn, hezh@sustech.edu.cn}
}
\maketitle

\begin{abstract}
A central challenge in human pose estimation, as well as in many other machine learning and prediction tasks, is the generalization problem. The learned network does not have the capability to characterize the prediction error, generate feedback information from the test sample, and correct the prediction error on the fly for each individual test sample, which results in degraded performance in generalization.
In this work, we introduce  a self-correctable and adaptable inference (SCAI) method to address the generalization challenge of network prediction and use human pose estimation as an example to demonstrate its effectiveness and performance.
We learn a correction network to correct the prediction result conditioned by a fitness feedback error. 
This feedback error is generated by a learned fitness feedback network which maps the prediction result to the original input domain and compares it against the original input. 
Interestingly, we find that this self-referential feedback error is highly correlated with the actual prediction error. 
This strong correlation suggests that we can use this error as feedback to guide the correction process. It can be also used as a loss function to quickly adapt and optimize the correction network during the inference process.
Our extensive experimental results on human pose estimation demonstrate that the proposed SCAI method is able to significantly
improve the generalization capability and performance of
human pose estimation.
\end{abstract}

\section{Introduction}
\label{sec:intro}
Human pose estimation (HPE) aims to correctly predict and localize human body joints. A variety of downstream applications are based on human pose estimation, such as motion capture \cite{DBLP:conf/cvpr/ElhayekAJTPABST15,DBLP:conf/cvpr/RhodinCKSF19}, activity recognition \cite{DBLP:conf/cvpr/BagautdinovAFFS17,DBLP:conf/cvpr/WuWWGW19, DBLP:conf/iccv/CheronLS15}, person tracking \cite{DBLP:conf/cvpr/YangRLZW021,DBLP:conf/cvpr/WangTM20} and video surveillance \cite{DBLP:conf/icmcs/LiC0H18}. Recently, deep learning-based methods for human pose estimation have achieved remarkable success \cite{DBLP:conf/cvpr/CaoSWS17,Chen_2018_CVPR,DBLP:conf/cvpr/0009XLW19,He_2017_ICCV,DBLP:conf/cvpr/PapandreouZKTTB17,DBLP:conf/cvpr/SuYXGW19}.  However, in complex or unseen scenarios, pose estimation remains very challenging due to occlusions, cluttered background, and large variations of appearance and scenes, especially for those distal keypoints at the end locations of body parts, such as wrists and ankles, which have large degrees of motion freedom and often suffer from severe occlusions \cite{DBLP:conf/eccv/XiaoWW18,DBLP:conf/cvpr/0005ZGH20}.

We recognize that one major challenge in current human
pose estimation, as well as in many other prediction tasks, is generalization. Network models, which have been well learned on the training set, often experience significant performance degradation on the test samples which are collected from different environments or scenarios. For example, in human pose estimation, there are different types of occlusions of body parts due to complex scene structures and free-style motions of human bodies.  More importantly, the occlusion scenarios of the test samples could be much different from those in the training samples.
This often leads to the significant performance degradation of human pose estimation from the training data to the test data.  For example, in our experiment, the average prediction accuracy on the training samples is 95.5\%. However, on the test set, this accuracy drops to 67\%. For those distal keypoints at tip locations of body parts which often experience more significant occlusions,  their average performance drop is much more significant, from 95.3\% to 57\%. 

To address this performance degradation or generalization problem, 
there are two major questions that need to be carefully answered: (1) how can we tell if the prediction is accurate or not during testing and how to characterize the prediction error? This is difficult because the ground truth values of the test samples are not available during testing.
Specifically, in pose estimation, we do not have the labeled ground truth locations of the body keypoints.
(2) How to correct the prediction error based on the specific characteristics of the test sample? 
Current network models, once successfully trained with labeled samples at the training side, remain fixed during testing, performing the feed-forward-only inference process to generate the prediction result. There is no mechanism for us to examine the specific characteristics of the test sample and use them as feedback to correct the prediction error or adjust the network model.
We believe that this unique capability of sample-specific prediction error characterization, error correction, and model optimization is very important for the generalization performance of learned network models. It also has the potential to significantly improve the prediction accuracy of test samples.

To address these two challenging issues, in this work, we propose to explore a learning-based feedback-control or correction method for prediction, with applications to human pose prediction.
Specifically, let $\mathbf{\hat{v}}=\mathbf{\Phi}(\mathbf{u})$ be the prediction network which is tasked to predict the true value of $\mathbf{v}$ from input $\mathbf{u}$. To answer the first question, we design and learn a fitness feedback network $\mathbf{\Gamma}$ 
which compares the  prediction result  $\mathbf{\hat{v}}=\mathbf{\Phi}(\mathbf{u})$ of the prediction network $\mathbf{\Phi}$ 
against the original input $\mathbf{u}$ and generate a self-referential  feedback error. Very interestingly, in this work, we find that this self-referential feedback error is highly correlated with the prediction error of the network $\mathbf{\Phi}$. Note that, when computing the self-referential error, we do not need the ground truth data. It can be directly computed on the input sample using the prediction-feedback networks. This allows us to characterize the prediction error of test samples. 

Under the guidance of self-referential error feedback, we train a prediction error correction network $\mathbf{C}$ to adjust the inference results during the prediction process to improve the prediction accuracy for the test samples. Besides, we find that the self-referential error and the fitness feedback network (FFN) can be used to construct a self-referential loss function on the test samples to quickly adapt and optimize the network model during the inference stage, making the model learnable on the test side. 
We apply the above self-correctable and adaptable inference (SCAI) method to human pose estimation.
Our extensive experimental results on benchmark datasets 
demonstrate that the proposed SCAI method is able to significantly improve the generalization capability of the underlying prediction algorithm. It outperforms the existing state-of-the-art methods on human pose estimation by large margins. For example, on the MS COCO-testdev dataset, our method improves upon the current best method by up to 1.4\%, which is quite significant.

\begin{figure*}[ht]
\centering
\includegraphics[width=1.7\columnwidth]{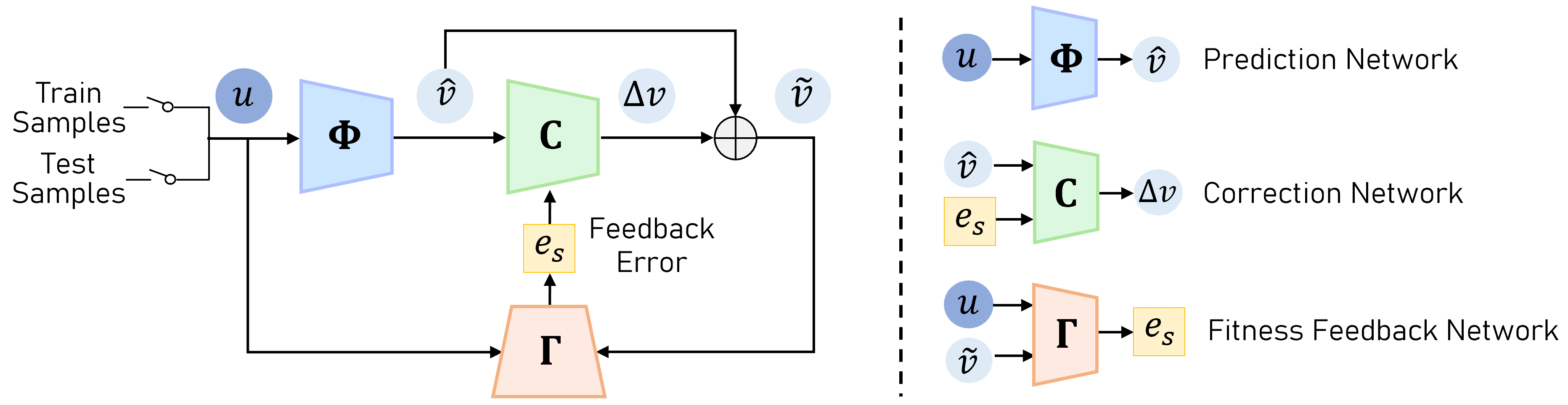}
\centering
\caption{An overview of our self-correctable and adaptable inference (SCAI) method.}
\label{fig:overview}
\end{figure*}

\section{Related Work}
\label{sec:relatedwork}
In this section, we review related works on human pose estimation, pose refinement, network generalization, and model adaptation. 

\textbf{(1) Human pose estimation.}  
There are two major categories of methods developed for human pose estimation: \textit{top-down}  and \textit{bottom-up} approaches. 
\textit{Top-down} methods \cite{DBLP:conf/cvpr/SuYXGW19,DBLP:conf/cvpr/0005ZGH20,Chen_2018_CVPR,DBLP:conf/cvpr/LiWZMFL19, DBLP:conf/eccv/SunXWLW18} first detect all persons in an image and then predict keypoint locations for each person. For example, Chen \etal \cite{Chen_2018_CVPR} utilized a human detector to generate bounding boxes as  the inputs into a feature pyramid network for  keypoint estimation. 
 \textit{Bottom-up} methods \cite{DBLP:conf/aaai/LiSW20,DBLP:conf/cvpr/LuoW00TZ21,DBLP:journals/corr/abs-2109-03622} first detect the joints for all persons in the input image and then group them into individual persons. For example,  Li \etal \cite{DBLP:conf/aaai/LiSW20} grouped detected keypoints into instances based on a greedy assignment algorithm. Luo \etal \cite{DBLP:conf/cvpr/LuoW00TZ21} introduced scale-adaptive  and weight-adaptive heatmap regression to alleviate large differences in human scales. 

\textbf{(2) Pose refinement.} A number of methods have been developed for pose refinement \cite{DBLP:conf/cvpr/CarreiraAFM16, DBLP:conf/cvpr/FieraruKPS18, DBLP:conf/cvpr/MoonCL19}. Fieraru \etal \cite{DBLP:conf/cvpr/FieraruKPS18} refined the estimation by exploiting dependencies between the input image and the body structure of the human pose through a network. Moon \etal \cite{DBLP:conf/cvpr/MoonCL19} trained a refinement network with synthetic poses generated from error distribution as input. Wang \etal \cite{DBLP:conf/eccv/WangLGDW20} extracted guidance keypoints from coarse pose estimation and applied a GCN as a refinement module to seek mutual information among keypoints. 
It should be noted that these refinement methods did not consider the sample distribution shift between the train set and test set and have not  addressed the generalization problem effectively.
Kan \etal \cite{DBLP:conf/eccv/KanCLH22} developed a prediction-verification network and perform a local search within the neighborhood of the prediction result. This multi-iteration search method is very time-consuming.

\textbf{(3) Generalization and adaptation.} 
Generalization is a significant challenge for existing network learning  methods. Test-time adaptation \cite{DBLP:conf/iclr/WangSLOD21, DBLP:conf/icml/SunWLMEH20} aims to improve generalization ability of the deep learning models when the test set exhibits a different data distribution from that of the train set. Sun \etal \cite{DBLP:conf/icml/SunWLMEH20} firstly trained self-supervised tasks on the test samples with model parameters optimized and then performed inference. 
Tung \etal \cite{DBLP:conf/nips/TungTYF17} performed test-time optimization with self-supervised losses driven by re-projection errors of keypoint, segmentation, and motion against detected 2D versions respectively. 
Li \etal \cite{DBLP:conf/nips/LiHDGW21} constructed a transformation between the self-supervised and the supervised keypoints, so the model can be fine-tuned toward the self-supervised objective of image reconstruction during inference. However, the self-supervised objectives of test-time adaptation utilized by these methods require specific conditions, such as projection from 3D to 2D or a set of images of the same person. Our method can predict human pose given general 2D images. Besides, our proposed self-referential error is highly correlated with the pose prediction error, which can be used to guide the prediction correction process and to construct self-referential loss for network model adaptation during test time.

\section{Method}
We first present our SCAI method  to address the generalization challenge in generic network learning and prediction. We then explain how this method can be applied to the specific problem of human pose estimation.

\subsection{Self-Correctable and Adaptable Inference}
In this section, we present the proposed self-correctable inference and adaptable inference method.

\textbf{(1) Overview}. As illustrated in Figure \ref{fig:overview}, let $\mathbf{\hat{v}}=\mathbf{\Phi(\mathbf{u})}$ be the prediction network which is tasked to predict the true value of $\mathbf{v}$ from input $u$. We learn a prediction error correction network $\mathbf{C}$ to produce a correction $\mathbf{\Delta v}$ which adjusts the prediction results $\mathbf{\hat{v}}$ to $\mathbf{\tilde{v}} = \mathbf{\hat{v}} + \mathbf{\Delta v}$, aiming to improve the prediction accuracy on the test samples. To guide such correction, we design an FFN $\mathbf{\Gamma}$ with the original input $\mathbf{u}$ and the corrected prediction $\mathbf{\tilde{v}}$ as inputs to generate a self-referential error. 
This error is used as important feedback to guide the correction network $\mathbf{C}$ to perform adaptive correction of the predicted result.
Besides, we find that the self-referential feedback error can be also used to define a loss function to quickly adapt and optimize the network model during the inference stage, making the model learnable on the test side. 

\textbf{(2) Self-Correctable Inference}. 
A deep neural network $\mathbf{\Phi}$, already being fully optimized on the training set, often suffers from significant performance degradation when applied to new or unseen test samples. The  prediction model remains fixed during testing, performing the feed-forward-only inference process to generate the vanilla prediction results $
   \mathbf{\hat{v}} = \mathbf{\Phi}(\mathbf{u})
$. As discussed above, we propose to learn a correction network $\mathbf{C}$ to correct the prediction error and improve the prediction results from  $\mathbf{\hat{v}}$ to $\mathbf{\tilde{v}}$ conditioned by an error feedback $\mathbf{e_s}$
\begin{equation}
   \mathbf{\Delta v} = \mathbf{C}(\mathbf{\hat{v}} | \mathbf{e_s}), \quad \mathbf{\tilde{v}} = \mathbf{\hat{v}} + \mathbf{\Delta v}.
\end{equation}
Note that the conditional error feedback  $\mathbf{e_s}$ is very important for the correction network. Without this feedback, the correction network cannot achieve any performance improvement since the original prediction network $\mathbf{\Phi}$ is already well-trained and fully optimized. Adding another feed-forward correction network simply does not help.

Now, the question is how can we generate a useful feedback signal $\mathbf{e_s}$ for the correction network $\mathbf{C}$?
To address this issue, we propose to introduce a fitness feedback network $\mathbf{\Gamma}$. It aims to 
evaluate how good the corrected prediction result is.
Note that we do not know the ground truth of the prediction result. Our idea is to map  the prediction result to the original input domain and compare it against the original input which has its ground truth value. Specifically, the fitness feedback network $\mathbf{\Gamma}$ takes two inputs, the corrected prediction $\mathbf{\tilde{v}}$ and the original input $\mathbf{u}$. The output of $\mathbf{\Gamma}$ is the so-called \textit{self-referential feedback error}
$\mathbf{e_s}$:
\begin{eqnarray}
\mathbf{e_s} &=& \mathbf{\Gamma}(\mathbf{\tilde{v}}, \mathbf{u}).
\end{eqnarray}
which is used to guide the correction network $\mathbf{C}$.
From Section 3.2, specifically, Figure \ref{fig:Correlation analysis}, we will see that, with successful training, $\mathbf{e_s}$ is highly correlated with the prediction error. This strong correlation allows us to use $\mathbf{e_s}$ as feedback to guide the correction of the prediction result. Otherwise, if the correction is weak, the correction process becomes unreliable and cannot achieve improved prediction performance.

\textbf{(3) Self-Adaptable Inference}. 
As its unique feature, the feedback error is self-referential. In other words, when computing this error, we do not need the ground truth value of the prediction. We only need the prediction network $\mathbf{\Phi}$, the fitness feedback network 
$\mathbf{\Gamma}$, the correction network $\mathbf{C}$, and the input. This implies that we can also compute this feedback error $\mathbf{e_s}$ during the network inference process. Once computed, its norm can be used as a loss function to quickly adapt and optimize the correction network $\mathbf{C}$   model during the inference stage using  gradient back-propagation.
During the update, the prediction network $\mathbf{\Phi}$ and the FFN $\mathbf{\Gamma}$ remain fixed.
It should be noted that this adaptation and optimization is only for the current test sample. Certainly, it can be extended to a batch or a cluster of test samples to reduce the complexity. 
In the following section, we will use human pose estimation as an example to explain the specific training procedures and loss function design for the proposed SCAI method.

\begin{figure*}[ht]
\centering
\includegraphics[width=2\columnwidth]{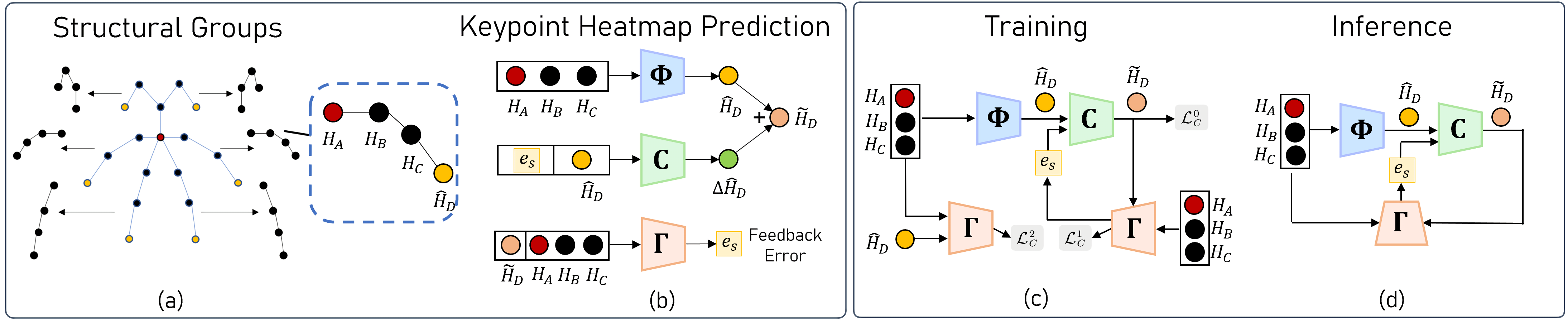}
\vspace{-5pt}
\caption{(a) Structural grouping of body keypoints. (b) Main network models. (c) Training process and (d) inference process of our SCAI method for human pose estimation. Only the refinement of the distal keypoints heatmap $\hat{H}_D$ is depicted as an example.}
\label{fig:correction_train}
\vspace{-10pt}
\end{figure*}

\subsection{SCAI for Human Pose Estimation}
\label{sec-pose}
In this work, we use human pose estimation as an example to implement the proposed SCAI method  and demonstrate its performance. 
The major question is: how should we re-structure the pose estimation problem so that we can apply the proposed prediction-feedback-correction scheme? 

\textbf{(1) Structural groups of body keypoints.}
Human pose estimation refers to the detection of body keypoint locations based on an input RGB image $I$ with dimensions $W \times H \times 3$. The task involves identifying $K$ keypoints $X=\{X_1,X_2, ...,X_K\}$ with high precision. Heatmap-based methods transform this problem into estimating $K$ heatmaps $\{H_1,H_2, ...,H_K\}$ of size $W' \times H'$. The location of a keypoint can be determined using different methods, such as grouping or peak finding, based on the corresponding heatmap \cite{DBLP:conf/eccv/XiaoWW18}. For instance, the pixel with the highest heatmap value can be selected as the keypoint location.

To enhance the generalization and prediction performance of human pose estimation using the proposed SCAI method, we adopt the approach introduced by Kan \etal \cite{DBLP:conf/eccv/KanCLH22} to partition the body keypoints into six structural groups, as depicted in Figure \ref{fig:correction_train}(a). Each structural group corresponds to a body part, with keypoints that are connected during motion. The group is further divided into a distal keypoint $X_D$ at the tip location of the body part, such as the wrists and ankles, and proximal keypoints ${X_A,X_B,X_C}$. We observe that distal keypoints often have larger errors in prediction due to more significant freedom of motion and possible occlusion by other objects. Our main objective is to leverage the proposed SCAI approach to improve the prediction accuracy and generalization capability of these distal keypoints.

\textbf{(2) SCAI Network Design.}
Given a set of keypoint heatmaps $\{H_A,H_B,H_C, H_D\}$ estimated by a baseline model, \eg HRNet \cite{DBLP:conf/cvpr/0009XLW19}, our task is to refine heatmaps $\{H_B,H_C, H_D\}$. Let us consider the refinement of heatmap $H_D$ of the distal keypoint $X_D$ as an example. The input to the prediction network $\mathbf{\Phi}$ is $\mathbf{u} = \{H_A,H_B,H_C\}$, the heatmaps for the proximal keypoints.
The prediction output is $\mathbf{\hat{v}} = \hat{H}_D$. The correction network $\mathbf{C}$ is used to refine the prediction result into $\mathbf{\tilde{v}} = \tilde{H}_D$. 
The output of  FFN $\mathbf{\Gamma}$ is the feedback error $\mathbf{e_s}$ to guide the correction. 

\begin{figure}[h]
\centering
\includegraphics[width=0.9\columnwidth, height=0.5\columnwidth]{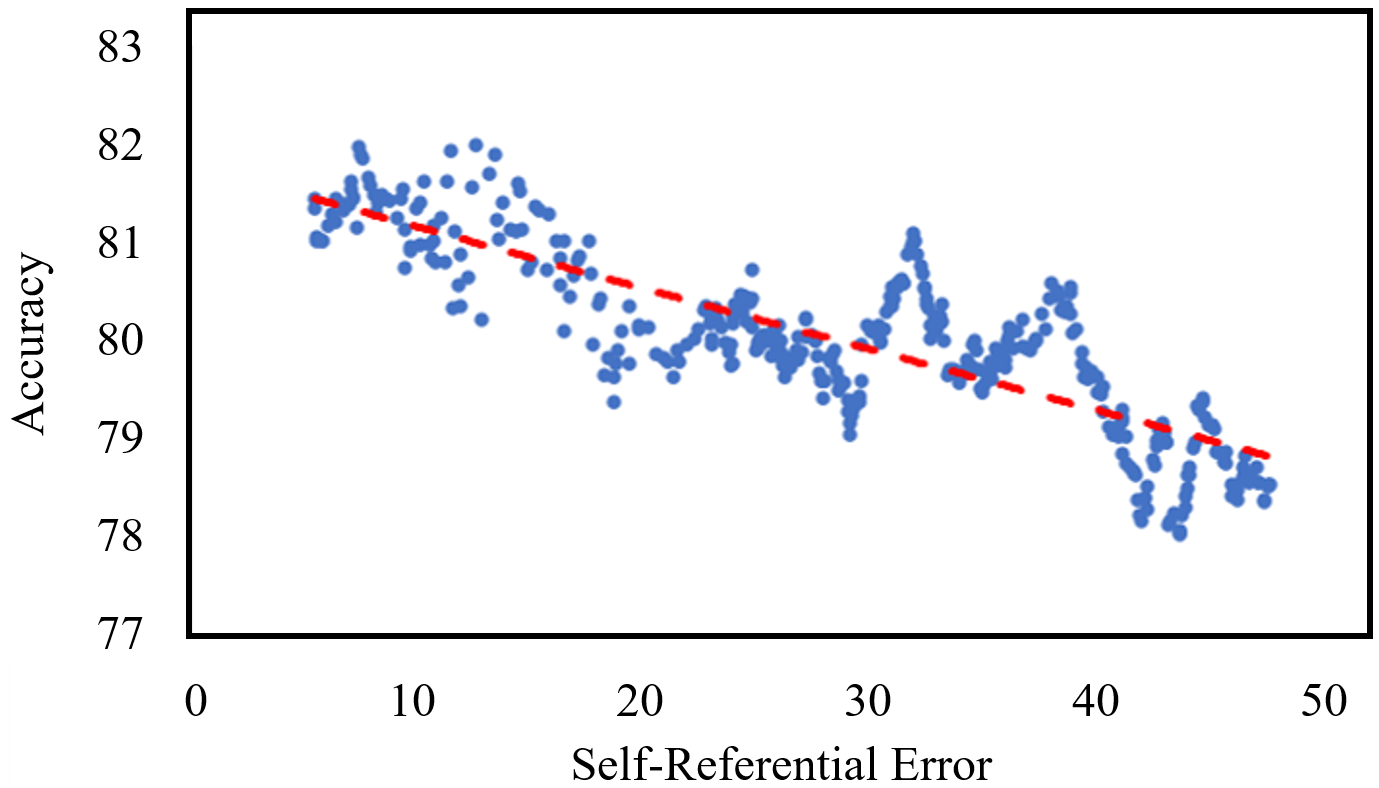}
\centering
\caption{Strong correlation between self-referential error and network prediction accuracy.}
\label{fig:Correlation analysis}
\vspace{-3pt}
\end{figure}

As illustrated in Figure \ref{fig:correction_train}(b), the prediction network $\mathbf{\Phi}$ predicts the distal keypoint heatmap $\hat{H}_D=\mathbf{\Phi}(H_A,H_B,H_C)$ using heatmaps of proximal keypoints $\{H_A,H_B,H_C\}$.
The correction network $\mathbf{C}$ aims to generate a corrected and improved estimation of $H_D$ conditioned by the feedback error $\mathbf{e_s}$
\begin{equation}
    \tilde{H}_D=\hat{H}_D + \mathbf{C}(\hat{H}_D | \mathbf{e_s}).
\end{equation}
Here, $\mathbf{e_s}$ is generated by the fitness feedback network  $\mathbf{\Gamma}$, which has two inputs, the corrected prediction result $\tilde{H}_D$ and the original inputs $[H_A,H_B,H_C]$:
\begin{equation}
\mathbf{e_s} = \mathbf{\Gamma}([H_A, H_B, H_C], \tilde{H}_D).
\end{equation}
In the following experiment, we demonstrate that the $L_2$-norm of self-referential feedback error  $\mathbf{e_s}$ is highly correlated with the actual network prediction error. In this human pose estimation experiment, we choose 400 batches of 25600 test samples to show this correlation. In Figure \ref{fig:Correlation analysis}, the vertical axis shows the average prediction accuracy of the body keypoints in each batch of test samples. The horizontal axis shows their average self-referential error. Each dot represents one test batch.
We can see that there is a very strong correlation between them, the corresponding correlation coefficient is $-0.84$.

\textbf{(3) SCAI Network Training.}
In the following, we explain how these three networks are trained with labeled samples.
Note that, at the training side, all the predicted keypoints have their ground truth values $\{H_A^{*},H_B^{*},H_C^{*}, H_D^{*}\}$. 
Therefore, for the prediction network $\mathbf{\Phi}$, its loss function is given by $\mathcal{L}_\Phi = \|\hat{H}_D - H_D^{*} \|_2$, which is the $L_2$ distance between the predicted heatmap and the ground truth value \cite{DBLP:conf/cvpr/0009XLW19}.
For the fitness feedback network $\mathbf{\Gamma}$, it
maps the prediction result $\tilde{H}_D$ to the original input domain $\{H_A, H_B, H_C\}$ and compare it against the original input. So, during the training stage, its loss function can be expressed in the form of self-referential error, given by  $\mathcal{L}_\Gamma = \|\hat{H}_A - H_A^{*} \|_2$.
For the correction network $\mathbf{C}$, its loss function is given by
\begin{eqnarray}
    \mathcal{L}_{C} = a\cdot\mathcal{L}_{C}^0 + b\cdot\mathcal{L}_C^1 + \lambda\cdot(\mathcal{L}_C^1 - \mathcal{L}_C^2),
\label{eq:correctionloss}
\end{eqnarray}
where $\mathcal{L}_{C}^0 = \|\tilde{H}_D - H_D^{*} \|_2$, $\mathcal{L}_{C}^1 = \|\hat{H}_A - H_A^{*} \|_2$ and $\mathcal{L}_{C}^2 = \|\bar{H}_A - H_A^{*} \|_2$, with $a$, $b$ and $\lambda$ representing the weights  for the three losses, respectively. 
$\mathcal{L}_C^1$ is defined by the distance between the mapping from the corrected prediction $\tilde{H}_D$ and the original input.
$\mathcal{L}_C^2$ is defined by the distance between the mapping from the uncorrected prediction $\hat{H}_D$ and the original input.
Here, the last term $\mathcal{L}_C^1 - \mathcal{L}_C^2$ is used to ensure the effectiveness of correction: the corrected result has a smaller self-referential error than the original prediction. Specifically, if the correction network improves the accuracy of keypoint prediction, the self-referential error evaluated by FFN will be smaller than the one without correction, that is, $\mathcal{L}_C^1$ value is smaller than $\mathcal{L}_C^2$.
Thus, the loss term $\mathcal{L}_C^1 - \mathcal{L}_C^2$ is able to guide the training of the correction network to effectively refine the keypoint prediction.

During the training process, the prediction network $\mathbf{\Phi}$ is pre-trained using the training samples
$\{[(H_A, H_B, H_C)\rightarrow H_D]\}$. The FFN is also pre-trained with training samples $\{[(H_B, H_C, H_D)\rightarrow H_A]\}$. 
Here, $\rightarrow$ represents the network prediction.
During the training stage, the prediction network $\mathbf{\Phi}$ is fixed. The pre-trained model of the FFN $\mathbf{\Gamma}$ is used as its initial model. The FFN and the correction networks are then jointly trained using their loss functions. Similar to the refinement of proximal keypoint ${X}_D$, we can also develop correction networks for keypoints ${X}_B$ and ${X}_C$ to optimize their heatmaps ${H}_B$ and ${H}_C$, respectively. More training details are shown in Supplemental Materials.

\begin{figure}[h]
\centering
\includegraphics[width=0.99\columnwidth]{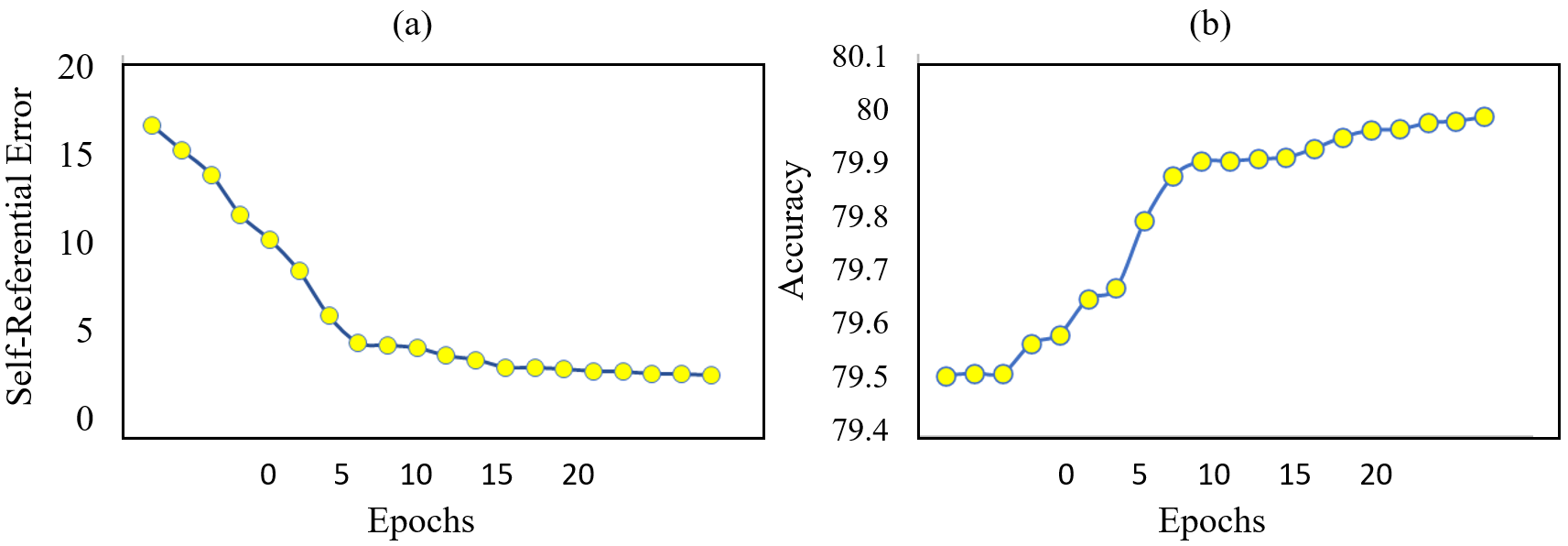}
\centering
\caption{The figure on the left is the loss decay curve of SCAI during training in the inference process, and the figure on the right is the test result corresponding to each epoch.}
\label{fig:test_time_sample}
\vspace{-10pt}
\end{figure}

\textbf{(4) Self-referential Adaptable Inference for Human Pose Estimation.}
Note that the self-referential error is computed based on the input samples and does not require the ground truth data of any body keypoints.
Therefore, on the test side, we use this self-referential error as a loss function to update the network model. Specifically, in this work, we choose to update the correction network $\mathbf{C}$ while other networks, including the prediction network $\mathbf{\Phi}$
and the FNN $\mathbf{\Gamma}$ remain fixed.
It should be noted that this model refinement is performed separately for each batch of test images. When moving to a new batch, the initial models obtained from the training set are restored and then refined. In other words, the models learned from one batch are not used for the next test batch to ensure flexible model adaptation. 
Figure \ref{fig:test_time_sample}(a) shows the  decreasing self-referential error and the convergence behavior of this model learning and adaptation process. Figure \ref{fig:test_time_sample}(b) shows that the accuracy of the test batch is consistently improved with the training epochs. 
This demonstrates that our self-referential adaptable inference method is able to use the test samples as feedback to update the network models and improve the prediction performance, providing natural and enhanced generalization capability for the learning and prediction network.

\begin{table*}[ht]
\centering
\setlength{\belowcaptionskip}{-0.5cm} 
\caption{
Comparison with the state-of-the-art methods on COCO test-dev. The best results are in \textbf{bold} and the second best results are \underline{underlined}.
}
\label{tab:sota on COCO}
\begin{tabular}{llccccccc}
\toprule
Method & Backbone & Size & {$AP$} & $AP^{50}$ & $AP^{75}$ &  $AP^{M}$ & $AP^{L}$ & $AR$  \\
\midrule
CFN \cite{DBLP:conf/iccv/HuangGT17} & -& -& 72.6& 86.1& 69.7& \textbf{78.3}& 64.1& -\\
CPN(ensemble) \cite{Chen_2018_CVPR}& ResNet-Incep. &384$\times$288 &73.0& 91.7& 80.9 &69.5& 78.1 &79.0\\
CSM+SCARB \cite{DBLP:conf/cvpr/SuYXGW19} & R152& 384$\times$288 &74.3 &91.8& 81.9 &70.7 &80.2 &80.5\\
CSANet \cite{DBLP:journals/corr/abs-1905-05355} & R152& 384$\times$288& 74.5& 91.7 &82.1& 71.2 &80.2& 80.7\\
HRNet \cite{DBLP:conf/cvpr/0009XLW19} & HR48& 384$\times$288 &75.5& 92.5& 83.3& 71.9 &81.5& 80.5\\
MSPN \cite{DBLP:journals/corr/abs-1901-00148} & MSPN &384$\times$288 &76.1& {93.4} &83.8& 72.3 &81.5& {81.6}\\
PoseFix \cite{DBLP:conf/cvpr/MoonCL19}&  HR48+R152& 384$\times$288 &76.7 &92.6& 84.1& 73.1& 82.6& 81.5\\
DARK \cite{DBLP:conf/cvpr/ZhangZD0Z20} &  HR48& 384$\times$288 &76.2& 92.5& 83.6& 72.5 &82.4& 81.1\\
UDP \cite{DBLP:conf/cvpr/0005ZGH20} &  HR48 &384$\times$288& 76.5 &{92.7}& 84.0& 73.0& 82.4& \underline{81.6}\\
Graph-PCNN \cite{DBLP:conf/eccv/WangLGDW20} & HR48 &384$\times$288 &{76.8}& 92.6& {84.3}& 73.3& {82.7} &\underline{81.6}\\
SCIO \cite{DBLP:conf/eccv/KanCLH22}
& HR48 &384$\times$288 &\underline{79.2}& \underline{93.5}& \underline{85.8}& 74.1& \underline{84.2} &\underline{81.6}\\ 
\midrule
\textbf{SCAI} (Ours) & HR48 & 384$\times$288 & \textbf{80.6} &  \textbf{94.8} & \textbf{87.0} & \underline{78.1} & \textbf{84.8} & \textbf{83.1}\\ 
\textbf{Performance Gain} & & & \textbf{+1.4} &\textbf{+1.3}&\textbf{+1.2}&\textbf{-0.2}&\textbf{+0.6}&\textbf{+1.5}\\
\bottomrule
\end{tabular}
\end{table*}

\section{Experiments}
\label{experiments}
This section presents a thorough evaluation of the proposed SCAI method, which includes experimental results, performance comparisons with state-of-the-art methods, and ablation studies. 

\subsection{Datasets}
In this section, we present the performance comparisons and ablation studies of our proposed SCAI method on two challenging datasets, namely the MS COCO \cite{DBLP:conf/eccv/LinMBHPRDZ14} and CrowdPose \cite{DBLP:conf/cvpr/LiWZMFL19}. The MS COCO dataset is a widely-used benchmark for human pose estimation, comprising 64K images with 270K annotated persons and 17 keypoints. The dataset includes diverse poses of multiple persons with varying body scales and occlusion patterns, making it challenging for human pose estimation models. Our models are trained on the train2017 split, which consists of 57K images with 150K persons, and ablation studies are conducted on the val2017 split. The CrowdPose dataset contains 20K images with 80K annotated persons and 14 keypoints. It includes crowded scenes and poses with various challenges. Unlike MS COCO, we partition the keypoints into 4 groups instead of 6 groups.  We train our models on the train set, which has 10K images and 35.4K persons and evaluate them on the validation set (2K images with 8K persons) and the test set (8K images with 29K persons).

\subsection{Experimental Settings}
For fair comparisons, we use the HRNet and ResNet as our baseline, from which pre-predicted results are produced to be further refined by our method. We follow the same training configuration in existing works \cite{DBLP:conf/eccv/XiaoWW18, DBLP:conf/cvpr/0009XLW19}. The correction network is trained with a full convolution network. We use the Adam \cite{kingma2014adam} optimizer for training. More experimental details are provided in the Supplemental Materials.

\begin{table}[h]
\centering
\resizebox{\linewidth}{!}{
\begin{threeparttable}
\caption{
Comparison with the state-of-the-art methods on CrowdPose test-dev. 
The best results are in \textbf{bold} and the second best results are \underline{underlined}.
}
\label{tab:sota on Crowdpose}
\begin{tabular}{lccc}
\toprule
Method & Backbone & $AP$  & $AP^{med}$  \\
\midrule
Mask-RCNN \cite{He_2017_ICCV}  & ResNet101 & 60.3  & -   \\
AlphaPose \cite{Fang_2017_ICCV}  &-     &61.0  &61.4\\
OccNet \cite{DBLP:conf/avss/GoldaKSB19}  & ResNet50 & 65.5  & 66.6 \\
JC-SPPE \cite{DBLP:conf/cvpr/LiWZMFL19}  & ResNet101 & 66.0  & 66.3    \\
HigherHRNet \cite{Cheng_2020_CVPR}  &HR48 & 67.6  &-  \\
MIPNet \cite{Khirodkar_2021_ICCV}  &HR48 & {70.0}  & {71.1}  \\
SCIO \cite{DBLP:conf/eccv/KanCLH22}  & HR48 & \underline{71.5}  & \underline{72.2}  \\
\midrule
\textbf{SCAI} (Ours) & HR48  & \textbf{72.4}  & \textbf{73.2} \\
\textbf{Performance Gain} &   & \textbf{+0.9}   & \textbf{+1.0}   \\
\bottomrule
\end{tabular}
\end{threeparttable}
}
\end{table}

\begin{table}[ht]
\centering
\resizebox{\linewidth}{!}{
\begin{threeparttable}
\caption{Comparison with state-of-the-art of three baselines on COCO test-dev.}
\label{tab:baseline}
\begin{tabular}{lccccccc}
\toprule
Method & Backbone  & $AP$ &  $AP^{M}$ & $AR$  \\
\midrule
SimpleBaseline & R152 & 73.7&70.3& 79.0\\
 + \textbf{SCAI}  & {R152}  & {\textbf{78.8}}  & {\textbf{75.5}} &  {\textbf{82.5}}\\
\textbf{Performance Gain} &      & \textbf{+5.1} & \textbf{+5.2}&  \textbf{+3.5}\\
\midrule

HRNet  & HR32&  74.9 & 71.3 & 80.1\\
+ \textbf{SCAI}  & HR32  & \textbf{79.9} & \textbf{76.8} &  \textbf{82.5}\\
\textbf{Performance Gain} &      & \textbf{+5.0} & \textbf{+5.5}&  \textbf{+2.4}\\
\midrule

HRNet  & HR48 &75.5& 71.9 & 80.5\\
+ \textbf{SCAI}  & HR48  & \textbf{80.6} & \textbf{78.1} &  \textbf{83.1}\\
\textbf{Performance Gain} &  & \textbf{+5.1} &\textbf{+6.2} & \textbf{+2.6}\\
\bottomrule
\end{tabular}
\end{threeparttable}
}
\end{table}

\begin{figure*}[h]
\centering
\includegraphics[width=1.85\columnwidth]{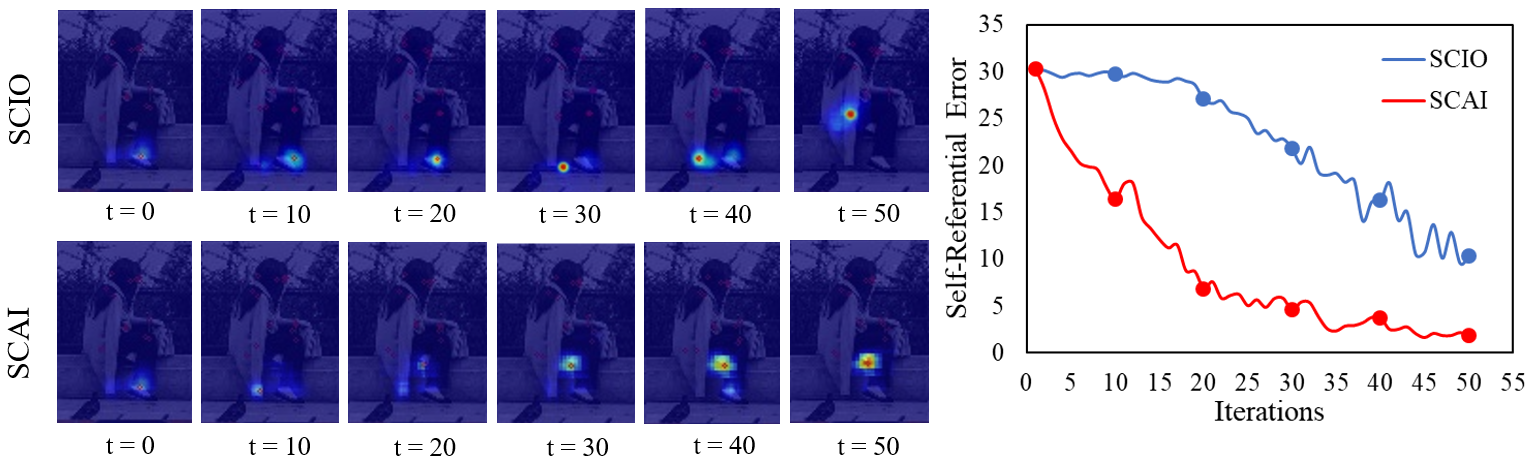}
\centering
\vspace{-10pt}
\caption{Comparison between the local search and the correction from SCIO and SCAI, where $t$ represents the number of iterations.}
\label{fig:discussion}
\vspace{-10pt}
\end{figure*}


\subsection{Performance Comparisons}
In Table \ref{tab:sota on COCO}, we compare the performance of our SCAI method  with the following state-of-the-art methods on the COCO test set: the 
CFN \cite{DBLP:conf/iccv/HuangGT17}, CPN (ensemble) \cite{Chen_2018_CVPR}, CSM+SCARB \cite{DBLP:conf/cvpr/SuYXGW19}, CSANet \cite{DBLP:journals/corr/abs-1905-05355}, HRNet \cite{DBLP:conf/cvpr/0009XLW19}, MSPN \cite{DBLP:journals/corr/abs-1901-00148}, PoseFix \cite{DBLP:conf/cvpr/MoonCL19}, DARK \cite{DBLP:conf/cvpr/ZhangZD0Z20}, UDP \cite{DBLP:conf/cvpr/0005ZGH20}, Graph-PCNN \cite{DBLP:conf/eccv/WangLGDW20} and SCIO \cite{DBLP:conf/eccv/KanCLH22} methods.
We can see that our SCAI method outperforms the current state-of-the-art methods by large margins, up to 1.4\%. 

Table \ref{tab:sota on Crowdpose} presents the results of our SCAI method compared to other state-of-the-art methods on the challenging CrowdPose dataset. The methods compared include Mask-RCNN \cite{He_2017_ICCV}, SimpleBaseline \cite{DBLP:conf/eccv/XiaoWW18}, AlphaPose \cite{Fang_2017_ICCV}, OccNet \cite{DBLP:conf/avss/GoldaKSB19}, JC-SPPE \cite{DBLP:conf/cvpr/LiWZMFL19}, HigherHRNet \cite{Cheng_2020_CVPR}, MIPNet \cite{Khirodkar_2021_ICCV}, and SCIO \cite{DBLP:conf/eccv/KanCLH22}. Our SCAI method achieves an improvement in average precision by up to 0.9\% compared to the current best method SCIO \cite{DBLP:conf/eccv/KanCLH22}. This indicates that our method can provide more accurate pose estimation for multi-person scenes with challenging occlusion scenarios.

Table \ref{tab:baseline} provides a comparison of our SCAI method with other state-of-the-art methods that use different baseline networks such as R152, HR32, and HR48 networks. Our SCAI method consistently outperforms other methods, even when different baseline networks are employed.

As Figure \ref{fig:discussion} shows, compared to the local search proposed in SCIO \cite{DBLP:conf/eccv/KanCLH22}, our method converges much faster and is able to stably correct the heatmap to be more accurate during the optimization process.

\subsection{Ablation Studies}
To assess the efficacy of our proposed SCAI method and analyze the influence of individual algorithmic components, we performed various ablation experiments on the COCO test-dev dataset. Our method incorporates two key innovations, namely self-correctable inference (SCI) and self-adaptable inference (SAI). Self-correctable inference consists of three essential parts: the correction network, self-referential error, and the joint training of the correction network and FFN. In Table \ref{tab:ablations}, we present the results of our ablation studies. The first row shows the performance of the baseline method. The second row indicates the performance when incorporating the correction network, while the third and fourth rows correspond to the addition of the self-referential error and joint training components, respectively. The last row reports the results when including the self-adaptable inference component. Our findings demonstrate that each algorithmic component makes a substantial contribution to the overall performance of our method.

\begin{table}[h] 
\centering
\setlength{\belowcaptionskip}{-0.3cm}
\setlength{\abovecaptionskip}{-0.03cm}
\caption{Ablation study on different algorithmic components on COCO test-dev dataset.}
\label{tab:ablations}
\resizebox{0.95\linewidth}{!}{
\begin{tabular}{lccc}
\toprule
 Method&    $AP$  &  $AP^{M}$  & $AR$   \\
\midrule
\textbf{Baseline} & 75.5&  71.9 & 80.5  \\
\midrule
\textbf{Baseline + SCI} & & &  \\
\,\, + \textit{Correction Network}& 78.5  & 73.9  &  81.4 \\
\,\, + \textit{Self-Referential Error} & 78.9  & 75.8  &  81.6 \\
\,\, + \textit{Joint Training} & 79.8  & 77.8  &  82.6 \\
\midrule
\textbf{Baseline + SCI + SAI} & & &  \\
\,\, + \textit{Self-Adaptive Inference} & \textbf{80.6} &  \textbf{78.1}  & \textbf{83.1}\\
\bottomrule
\end{tabular}
}
\centering
\vspace{-5pt}
\end{table}



Table \ref{tab:weight} presents an analysis of the impact of varying loss weights in Equation (\ref{eq:correctionloss}) on our experimental results. The results suggest that the optimal weights for the parameters $a$, $b$, and $\lambda$ are 0.85, 0.65, and 0.45, respectively.

\begin{table}[h]
\vspace{-7pt}
\centering
\setlength{\abovecaptionskip}{-0.02cm}
\caption{Ablation study on different loss parts and their weights in Equation (\ref{eq:correctionloss}).} 
\label{tab:weight}

\begin{tabular}{ccc|ccc}
\toprule
 $a$  &  $b$  &  $\lambda$  &  $AP$  &  $AP^{M}$  & $AR$   \\
\midrule
0.85 & 0.65 & 0.30 &80.2 &  77.6 & 82.7  \\
0.85 & 0.65 & 0.60 &80.4&  77.9 & 82.9  \\
0.85 & 0.65 & 0.45 & \textbf{80.6} &  \textbf{78.1}  & \textbf{83.1}\\
1.00 & 0.65 & 0.45 &80.5&  78.1 & 83.1  \\
0.85 & 0.40 & 0.45 & 80.4 &  77.9  & 83.1\\
\bottomrule
\end{tabular}
\centering
\end{table}

\begin{figure}[ht]
\centering
\includegraphics[width=\columnwidth]{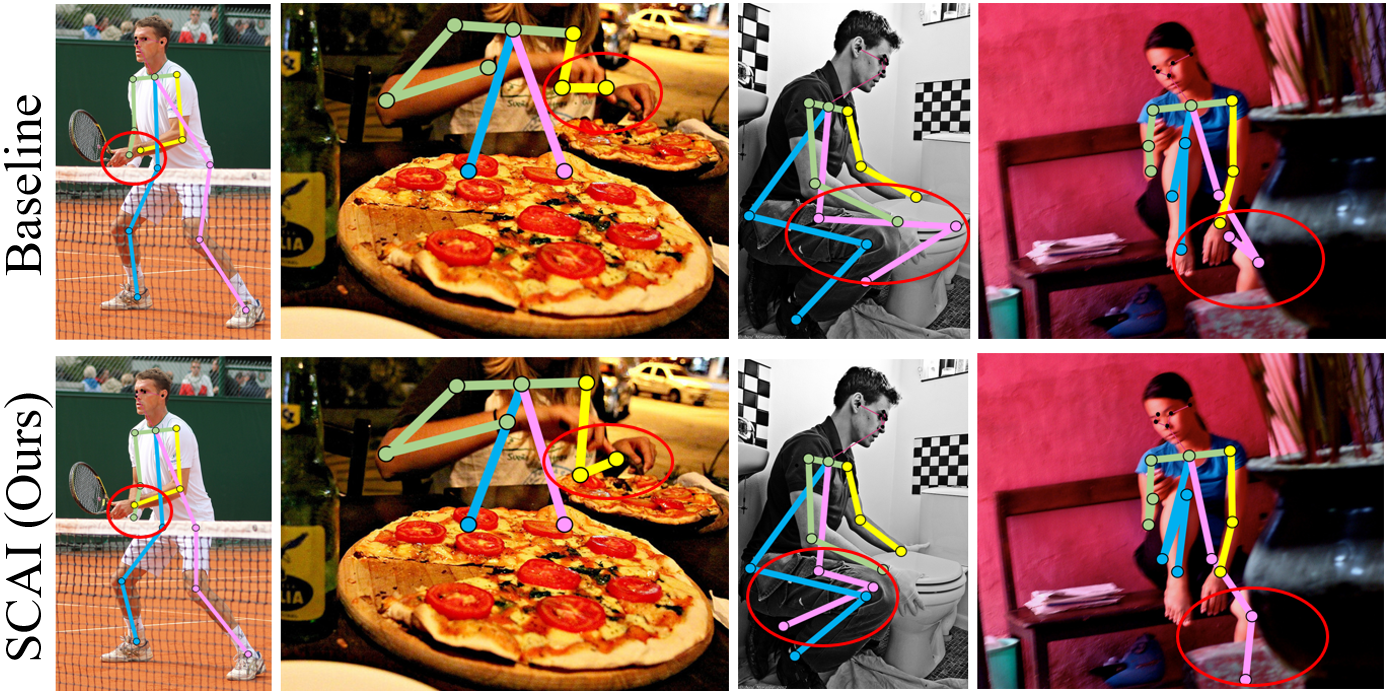}
\centering
\caption{Four examples of refinement of predicted keypoints. The top row is the original estimation. The bottom row is the estimation from SCAI.}
\label{fig:sample1}
\vspace{-5pt}
\end{figure}

Figure \ref{fig:sample1} presents a comparison of the pose estimation results refined by our proposed SCAI method against those produced by the baseline method. To this end, we randomly selected four images from the COCO val2017 dataset. The pose estimation results from the baseline method are displayed in the top row. In the bottom row, we show the refined keypoint estimations produced by our SCAI method. As can be observed, our method successfully corrects the keypoints circled in red, such as the left knee of the person in the third column. This indicates the effectiveness of our approach in refining pose estimation results.

\begin{figure}[ht]
\setlength{\abovecaptionskip}{-0.02cm}
\centering
\includegraphics[width=\columnwidth]{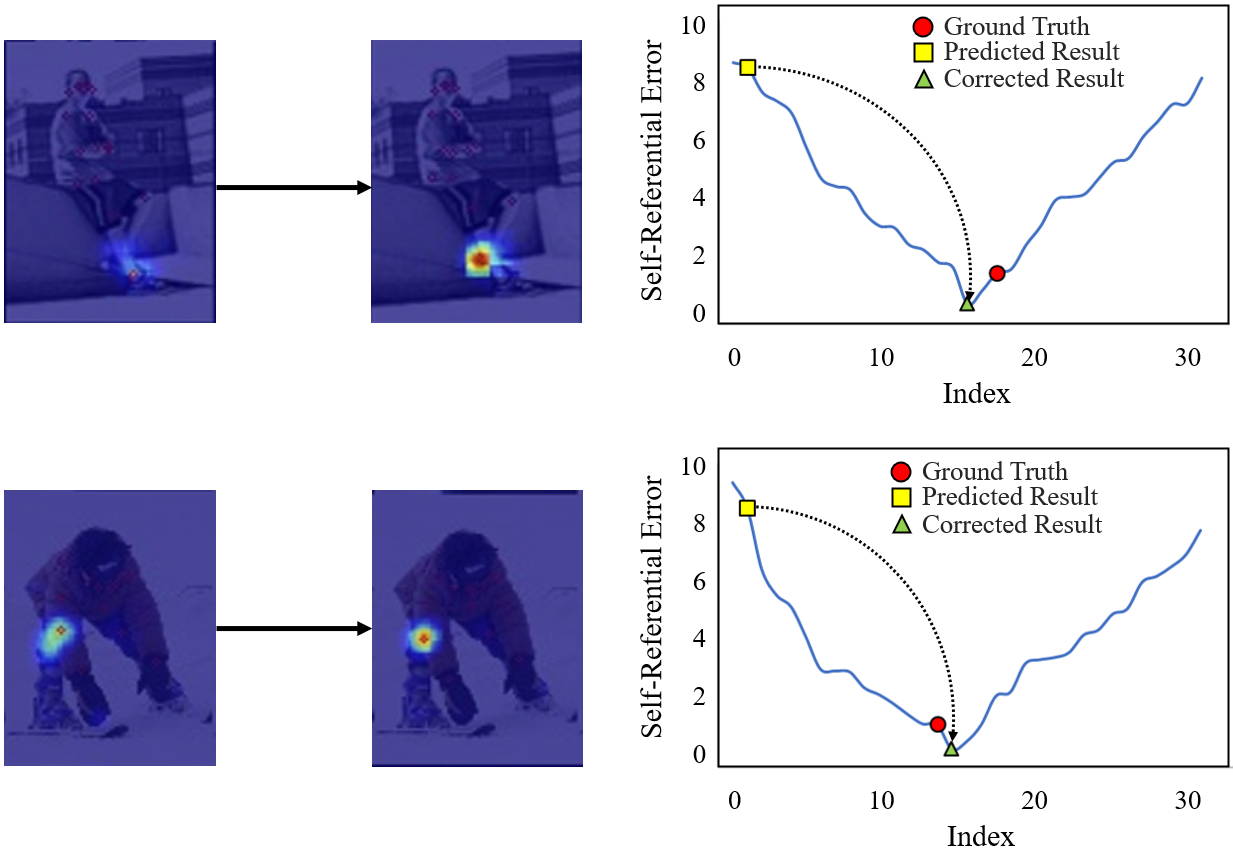}
\centering
\caption{Two examples of corrected keypoints from SCAI. The distribution shows keypoints error by the ground truth (red dot), prediction network (yellow square) and our method SCAI (green triangle),
where the blue curves represent errors of randomly selected from correction process. }
\label{fig:correction_sample}
\vspace{-10pt}
\end{figure}

In our self-correctable  network design, we aim to correct the prediction output so as to minimize the self-referential error. 
Figure \ref{fig:correction_sample} shows two examples where the self-referential error can be used as a feedback reference to guide the correction network to pull the prediction result toward the point with the minimum self-referential error. More importantly, this point is often very close to the ground truth (circle) since the self-referential feedback error is highly correlated with the keypoint prediction error.

\section{Further Discussions and Summary of Unique Contributions}
In this section, we provide further discussion of our proposed SCAI method compared with related work
and summarize its unique contributions.

\textbf{(1) Unique Differences from Related Work.} 
This work is related to the SCIO \cite{DBLP:conf/eccv/KanCLH22}. In our design, we borrowed the structural grouping of the human body keypoints from the paper. We also used its verification network to pre-train our fitness feedback network. 
Compared to this paper, the main part of our algorithm, namely, the self-correctable and adaptable inference method, including the correction network, fitness feedback network, and inference-time network optimization, are totally new. From the experimental results, we can see that our SCAI algorithm outperforms the method in Kan \etal \cite{DBLP:conf/eccv/KanCLH22} by large margins. Kan \etal \cite{DBLP:conf/eccv/KanCLH22} performed a local search to refine the prediction result with very high computational complexity. In this work, we learned a correction network to correct the prediction error which is much more effective and has much lower computational complexity. Also, as mentioned in section 4.3, our method performs more accurate keypoint estimation on the optimization process during the inference stage.

This work is also related to cycle consistency \cite{Zhu_2017_ICCV} , reciprocal learning \cite{Sun_2020_CVPR} and dual learning \cite{DBLP:conf/nips/HeXQWYLM16,DBLP:conf/icml/XiaQCBYL17}. 
Compared to these methods, our method is uniquely different and novel since our method establishes the correctable and adaptable inference with the ability to correct the prediction result and update the network model during the inference stage. However, such correction processes and adaptable inference are not available in the above existing methods and they only used cycle constraints for model training and testing.

\textbf{(2) Algorithm complexity.}
Table \ref{tab:ablations with Decision Stack} demonstrates that our proposed SCAI method introduces additional computational complexity. Specifically, the FFN and correction network increase the complexity of the baseline pose estimation model. It is worth noting that these two networks operate on the previously predicted heatmaps and thus their network structures and complexity are relatively modest. The self-referential adaptation requires updating the correction network multiple times for the entire batch, thereby increasing the overall computational burden. However, compared to SCIO, our model offers improved inference speed. In our future work, we aim to explore more efficient methods of self-adaptable inference.

\begin{table}[ht]
\begin{center}
\vspace{-5pt}
\caption{Complexity analysis on COCO val set.} 
\vspace{-10pt}
\label{tab:ablations with Decision Stack}
\begin{tabular}{l|cc}
\toprule
 Method &  Parameters (M) &  Speed (fps) \\
\midrule
HRNet \cite{DBLP:conf/cvpr/0009XLW19}             & \textbf{64} & \textbf{125}\\
SCIO \cite{DBLP:conf/eccv/KanCLH22}      & 193 & 72\\
SCAI (Ours)      & 357 & 81\\
\bottomrule
\end{tabular}
\end{center}
\vspace{-10pt}
\end{table}

\textbf{(3) Summary of contributions.} 
The major contributions of this work can be summarized as follows: 
(a) We demonstrate that it is theoretically possible to learn a feedback-correction network to refine the prediction results of a well-trained network, outperforming the SOTA and SCIO by 1.4\%, which is quite significant. 
(b) We have introduced a correction network which is able to correct the prediction error for the test sample guided by the self-referential feedback error. This error was generated by a learned fitness feedback network. We found that this self-referential error is highly correlated with the actual network prediction error. 
(c) Using the self-referential error, we have introduced a new loss function to perform quick adaptation and optimization of the correction network during the inference stage.
(d) We apply the proposed self-correctable and adaptable inference method to human pose estimation and have achieved remarkable performance gain and significant improvement of generalization capability of the pose estimation network.

\section{Conclusion}
In this work, we have developed  a self-correctable and adaptable inference method to address the generalization challenge of network prediction and use human pose estimation as an example to demonstrate its effectiveness and performance.
We have introduced the self-referential feedback error on the network input samples by constructing a feedback fitness network, so that it is able to evaluate if the prediction is accurate or not, without the need to know the ground truth. Guided by the self-referential error,  we learn a prediction correction network which is able to adjust the prediction result during the dynamic inference process. 
The self-referential error is employed as a loss function for network adaptation during the inference phase. Our comprehensive experimental results on human pose estimation attest to the ability of the SCAI method to enhance both the generalization ability and performance of human pose estimation to a significant degree.


{\small
\bibliographystyle{ieee_fullname}
\bibliography{egbib}

\begin{thebibliography}{10}\itemsep=-1pt

\bibitem{DBLP:conf/cvpr/BagautdinovAFFS17}
Timur~M. Bagautdinov, Alexandre Alahi, Fran{\c{c}}ois Fleuret, Pascal Fua, and
  Silvio Savarese.
\newblock Social scene understanding: End-to-end multi-person action
  localization and collective activity recognition.
\newblock In {\em {CVPR}}, pages 3425--3434, 2017.

\bibitem{DBLP:conf/cvpr/CaoSWS17}
Zhe Cao, Tomas Simon, Shih{-}En Wei, and Yaser Sheikh.
\newblock Realtime multi-person 2d pose estimation using part affinity fields.
\newblock In {\em CVPR}, pages 1302--1310, 2017.

\bibitem{DBLP:conf/cvpr/CarreiraAFM16}
Joao Carreira, Pulkit Agrawal, Katerina Fragkiadaki, and Jitendra Malik.
\newblock Human pose estimation with iterative error feedback.
\newblock In {\em CVPR}, pages 4733--4742, 2016.

\bibitem{Chen_2018_CVPR}
Yilun Chen, Zhicheng Wang, Yuxiang Peng, Zhiqiang Zhang, Gang Yu, and Jian Sun.
\newblock Cascaded pyramid network for multi-person pose estimation.
\newblock In {\em CVPR}, pages 7103--7112, 2018.

\bibitem{Cheng_2020_CVPR}
Bowen Cheng, Bin Xiao, Jingdong Wang, Honghui Shi, Thomas~S. Huang, and Lei
  Zhang.
\newblock Higherhrnet: Scale-aware representation learning for bottom-up human
  pose estimation.
\newblock In {\em CVPR}, pages 5385--5394, 2020.

\bibitem{DBLP:conf/iccv/CheronLS15}
Guilhem Ch{\'e}ron, Ivan Laptev, and Cordelia Schmid.
\newblock P-cnn: Pose-based cnn features for action recognition.
\newblock In {\em ICCV}, pages 3218--3226, 2015.

\bibitem{DBLP:conf/cvpr/ElhayekAJTPABST15}
Ahmed Elhayek, Edilson de Aguiar, Arjun Jain, Jonathan Tompson, Leonid
  Pishchulin, Mykhaylo Andriluka, Christoph Bregler, Bernt Schiele, and
  Christian Theobalt.
\newblock Efficient convnet-based marker-less motion capture in general scenes
  with a low number of cameras.
\newblock In {\em CVPR}, pages 3810--3818, 2015.

\bibitem{Fang_2017_ICCV}
Hao-Shu Fang, Shuqin Xie, Yu-Wing Tai, and Cewu Lu.
\newblock Rmpe: Regional multi-person pose estimation.
\newblock In {\em ICCV}, pages 2334--2343, 2017.

\bibitem{DBLP:conf/cvpr/FieraruKPS18}
Mihai Fieraru, Anna Khoreva, Leonid Pishchulin, and Bernt Schiele.
\newblock Learning to refine human pose estimation.
\newblock In {\em CVPRW}, pages 205--214, 2018.

\bibitem{DBLP:conf/avss/GoldaKSB19}
Thomas Golda, Tobias Kalb, Arne Schumann, and J{\"u}rgen Beyerer.
\newblock Human pose estimation for real-world crowded scenarios.
\newblock In {\em AVSS}, pages 1--8. IEEE, 2019.

\bibitem{DBLP:conf/nips/HeXQWYLM16}
Di He, Yingce Xia, Tao Qin, Liwei Wang, Nenghai Yu, Tie{-}Yan Liu, and
  Wei{-}Ying Ma.
\newblock Dual learning for machine translation.
\newblock In {\em {NeurIPS}}, pages 820--828, 2016.

\bibitem{He_2017_ICCV}
Kaiming He, Georgia Gkioxari, Piotr Doll{\'a}r, and Ross Girshick.
\newblock Mask r-cnn.
\newblock In {\em ICCV}, pages 2961--2969, 2017.

\bibitem{DBLP:conf/cvpr/0005ZGH20}
Junjie Huang, Zheng Zhu, Feng Guo, and Guan Huang.
\newblock The devil is in the details: Delving into unbiased data processing
  for human pose estimation.
\newblock In {\em {CVPR}}, pages 5699--5708, 2020.

\bibitem{DBLP:conf/iccv/HuangGT17}
Shaoli Huang, Mingming Gong, and Dacheng Tao.
\newblock A coarse-fine network for keypoint localization.
\newblock In {\em {ICCV}}, pages 3047--3056, 2017.

\bibitem{DBLP:conf/eccv/KanCLH22}
Zhehan Kan, Shuoshuo Chen, Zeng Li, and Zhihai He.
\newblock Self-constrained inference optimization on structural groups for
  human pose estimation.
\newblock In {\em ECCV}, volume 13665, pages 729--745. Springer, 2022.

\bibitem{Khirodkar_2021_ICCV}
Rawal Khirodkar, Visesh Chari, Amit Agrawal, and Ambrish Tyagi.
\newblock Multi-instance pose networks: Rethinking top-down pose estimation.
\newblock In {\em ICCV}, pages 3122--3131, October 2021.

\bibitem{kingma2014adam}
Diederik~P Kingma and Jimmy Ba.
\newblock Adam: A method for stochastic optimization.
\newblock {\em arXiv preprint arXiv:1412.6980}, 2014.

\bibitem{DBLP:conf/icmcs/LiC0H18}
Dangwei Li, Xiaotang Chen, Zhang Zhang, and Kaiqi Huang.
\newblock Pose guided deep model for pedestrian attribute recognition in
  surveillance scenarios.
\newblock In {\em ICME}, pages 1--6. IEEE, 2018.

\bibitem{DBLP:conf/aaai/LiSW20}
Jia Li, Wen Su, and Zengfu Wang.
\newblock Simple pose: Rethinking and improving a bottom-up approach for
  multi-person pose estimation.
\newblock In {\em AAAI}, volume~34, pages 11354--11361, 2020.

\bibitem{DBLP:conf/cvpr/LiWZMFL19}
Jiefeng Li, Can Wang, Hao Zhu, Yihuan Mao, Hao-Shu Fang, and Cewu Lu.
\newblock Crowdpose: Efficient crowded scenes pose estimation and a new
  benchmark.
\newblock In {\em CVPR}, pages 10863--10872, 2019.

\bibitem{DBLP:journals/corr/abs-1901-00148}
Wenbo Li, Zhicheng Wang, Binyi Yin, Qixiang Peng, Yuming Du, Tianzi Xiao, Gang
  Yu, Hongtao Lu, Yichen Wei, and Jian Sun.
\newblock Rethinking on multi-stage networks for human pose estimation.
\newblock {\em arXiv preprint arXiv:1901.00148}, 2019.

\bibitem{DBLP:conf/nips/LiHDGW21}
Yizhuo Li, Miao Hao, Zonglin Di, Nitesh~B. Gundavarapu, and Xiaolong Wang.
\newblock Test-time personalization with a transformer for human pose
  estimation.
\newblock In {\em NeurIPS}, pages 2583--2597, 2021.

\bibitem{DBLP:conf/eccv/LinMBHPRDZ14}
Tsung-Yi Lin, Michael Maire, Serge Belongie, James Hays, Pietro Perona, Deva
  Ramanan, Piotr Doll{\'a}r, and C~Lawrence Zitnick.
\newblock Microsoft coco: Common objects in context.
\newblock In {\em ECCV}, pages 740--755. Springer, 2014.

\bibitem{DBLP:conf/cvpr/LuoW00TZ21}
Zhengxiong Luo, Zhicheng Wang, Yan Huang, Liang Wang, Tieniu Tan, and Erjin
  Zhou.
\newblock Rethinking the heatmap regression for bottom-up human pose
  estimation.
\newblock In {\em CVPR}, pages 13264--13273, 2021.

\bibitem{DBLP:conf/cvpr/MoonCL19}
Gyeongsik Moon, Ju~Yong Chang, and Kyoung~Mu Lee.
\newblock Posefix: Model-agnostic general human pose refinement network.
\newblock In {\em CVPR}, pages 7773--7781, 2019.

\bibitem{DBLP:conf/cvpr/PapandreouZKTTB17}
George Papandreou, Tyler Zhu, Nori Kanazawa, Alexander Toshev, Jonathan
  Tompson, Chris Bregler, and Kevin Murphy.
\newblock Towards accurate multi-person pose estimation in the wild.
\newblock In {\em CVPR}, pages 3711--3719, 2017.

\bibitem{DBLP:conf/cvpr/RhodinCKSF19}
Helge Rhodin, Victor Constantin, Isinsu Katircioglu, Mathieu Salzmann, and
  Pascal Fua.
\newblock Neural scene decomposition for multi-person motion capture.
\newblock In {\em {CVPR}}, pages 7703--7713, 2019.

\bibitem{DBLP:conf/cvpr/SuYXGW19}
Kai Su, Dongdong Yu, Zhenqi Xu, Xin Geng, and Changhu Wang.
\newblock Multi-person pose estimation with enhanced channel-wise and spatial
  information.
\newblock In {\em {CVPR}}, pages 5674--5682. Computer Vision Foundation /
  {IEEE}, 2019.

\bibitem{Sun_2020_CVPR}
Hao Sun, Zhiqun Zhao, and Zhihai He.
\newblock Reciprocal learning networks for human trajectory prediction.
\newblock In {\em CVPR}, pages 7414--7423, 2020.

\bibitem{DBLP:conf/cvpr/0009XLW19}
Ke Sun, Bin Xiao, Dong Liu, and Jingdong Wang.
\newblock Deep high-resolution representation learning for human pose
  estimation.
\newblock In {\em CVPR}, pages 5693--5703, 2019.

\bibitem{DBLP:conf/eccv/SunXWLW18}
Xiao Sun, Bin Xiao, Fangyin Wei, Shuang Liang, and Yichen Wei.
\newblock Integral human pose regression.
\newblock In {\em {ECCV}}, pages 536--553, 2018.

\bibitem{DBLP:conf/icml/SunWLMEH20}
Yu Sun, Xiaolong Wang, Zhuang Liu, John Miller, Alexei~A. Efros, and Moritz
  Hardt.
\newblock Test-time training with self-supervision for generalization under
  distribution shifts.
\newblock In {\em {ICML}}, volume 119, pages 9229--9248. {PMLR}, 2020.

\bibitem{DBLP:conf/nips/TungTYF17}
Hsiao{-}Yu Tung, Hsiao{-}Wei Tung, Ersin Yumer, and Katerina Fragkiadaki.
\newblock Self-supervised learning of motion capture.
\newblock In {\em {NeurIPS}}, pages 5236--5246, 2017.

\bibitem{DBLP:conf/iclr/WangSLOD21}
Dequan Wang, Evan Shelhamer, Shaoteng Liu, Bruno~A. Olshausen, and Trevor
  Darrell.
\newblock Tent: Fully test-time adaptation by entropy minimization.
\newblock In {\em {ICLR}}. OpenReview.net, 2021.

\bibitem{DBLP:conf/eccv/WangLGDW20}
Jian Wang, Xiang Long, Yuan Gao, Errui Ding, and Shilei Wen.
\newblock Graph-pcnn: Two stage human pose estimation with graph pose
  refinement.
\newblock In {\em ECCV}, pages 492--508. Springer, 2020.

\bibitem{DBLP:conf/cvpr/WangTM20}
Manchen Wang, Joseph Tighe, and Davide Modolo.
\newblock Combining detection and tracking for human pose estimation in videos.
\newblock In {\em {CVPR}}, pages 11085--11093, 2020.

\bibitem{DBLP:conf/cvpr/WuWWGW19}
Jianchao Wu, Limin Wang, Li Wang, Jie Guo, and Gangshan Wu.
\newblock Learning actor relation graphs for group activity recognition.
\newblock In {\em {CVPR}}, pages 9964--9974, 2019.

\bibitem{DBLP:conf/icml/XiaQCBYL17}
Yingce Xia, Tao Qin, Wei Chen, Jiang Bian, Nenghai Yu, and Tie{-}Yan Liu.
\newblock Dual supervised learning.
\newblock In {\em {ICML}}, volume~70 of {\em Proceedings of Machine Learning
  Research}, pages 3789--3798. {PMLR}, 2017.

\bibitem{DBLP:conf/eccv/XiaoWW18}
Bin Xiao, Haiping Wu, and Yichen Wei.
\newblock Simple baselines for human pose estimation and tracking.
\newblock In {\em ECCV}, pages 466--481, 2018.

\bibitem{DBLP:journals/corr/abs-2109-03622}
Nan Xue, Tianfu Wu, Gui-Song Xia, and Liangpei Zhang.
\newblock Learning local-global contextual adaptation for multi-person pose
  estimation.
\newblock In {\em CVPR}, pages 13065--13074, 2022.

\bibitem{DBLP:conf/cvpr/YangRLZW021}
Yiding Yang, Zhou Ren, Haoxiang Li, Chunluan Zhou, Xinchao Wang, and Gang Hua.
\newblock Learning dynamics via graph neural networks for human pose estimation
  and tracking.
\newblock In {\em {CVPR}}, pages 8074--8084, 2021.

\bibitem{DBLP:journals/corr/abs-1905-05355}
Dongdong Yu, Kai Su, Xin Geng, and Changhu Wang.
\newblock A context-and-spatial aware network for multi-person pose estimation.
\newblock {\em arXiv preprint arXiv:1905.05355}, 2019.

\bibitem{DBLP:conf/cvpr/ZhangZD0Z20}
Feng Zhang, Xiatian Zhu, Hanbin Dai, Mao Ye, and Ce Zhu.
\newblock Distribution-aware coordinate representation for human pose
  estimation.
\newblock In {\em CVPR}, pages 7091--7100, 2020.

\bibitem{Zhu_2017_ICCV}
Jun-Yan Zhu, Taesung Park, Phillip Isola, and Alexei~A. Efros.
\newblock Unpaired image-to-image translation using cycle-consistent
  adversarial networks.
\newblock In {\em ICCV}, pages 2242--2251, 2017.

\end{thebibliography}
}

\end{document}